\documentclass[11pt]{article}
\usepackage{coling2018}
\usepackage{times}
\usepackage{url}
\usepackage{latexsym}
\usepackage{blindtext}
\usepackage{graphicx}
\usepackage{subcaption}
\usepackage{xcolor}

\title{Benchmarking Neural Machine Translation for Southern African Languages}

\author{Laura Martinus \\
  Explore Data Science Academy, South Africa \\
  {\tt laura@explore-ai.net} \\\And
  Jade Z. Abbott \\
  Retro Rabbit, South Africa  \\
  {\tt jabbott@retrorabbit.co.za} \\}

\date{}

\begin{document}
\maketitle

\begin{abstract}

Unlike major Western languages, most African languages are very low-resourced. Furthermore, the resources that do exist are often scattered and difficult to obtain and discover. As a result, the data and code for existing research has rarely been shared. This has lead a struggle to reproduce reported results, and few publicly available benchmarks for African machine translation models exist. To start to address these problems, we trained neural machine translation models for 5 Southern African languages on publicly-available datasets. Code is provided for training the models and evaluate the models on a newly released evaluation set, with the aim of spur future research in the field for Southern African languages.

\end{abstract}

\section{Introduction}
Africa has over 2 000 languages across the continent \cite{enthnologue}. South Africa itself has 11 official languages, and many more unofficial languages. Unlike major Western languages, the multitude of African languages are very low-resourced. Furthermore, the resources that do exist are often scattered and difficult to obtain or discover. The datasets are often owned by small institutions and research is published in smaller African conferences and never published online. As a result, the data and code for existing research has rarely been shared, meaning researchers struggle to reproduce the results properly, and few publicly available benchmarks for African machine translation models exist.

To begin to solve the highlighted problems, we train neural machine translation (NMT) models for a subset of Southern African languages, for which parallel publicly-available datasets exist and publish the code on GitHub\footnote{Available online at: \url{https://github.com/LauraMartinus/ukuxhumana}}. We evaluate the models on the new Autshumato evaluation set for machine translation of Southern African languages \cite{mckellar2017autshumato}, in an attempt to spur future research in machine translation of Southern African languages.

Translation models are trained for English to Afrikaans, isiZulu, Northern Sotho, Setswana and Xitsonga, using state-of-the-art NMT architectures, namely, Convolutional Sequence-to-Sequence (ConvS2S) and Transformer architectures \cite{chahuneau2013translating}. 

\blfootnote{\hspace{-0.65cm}\vspace{-0.65cm} Presented at ACL 2019 Workshop on Widening Natural Language Processing}

\section{Languages}
\label{languages}

The isiZulu, N. Sotho, Setswana, and Xitsonga languages belong to the Southern Bantu group of African languages \cite{mesthrie2002language}. The Bantu languages are agglutinative and all exhibit a rich noun class system, subject-verb-object word order, and tone \cite{zerbian2007first}. Northern Sotho and Setswana are closely related and highly mutually-intelligible. Xitsonga is a language of the Vatsonga people \cite{100yearsBill}. The language of isiZulu belongs to the Nguni language family, and is known for its morphological complexity \cite{keet2017grammar}. Afrikaans is an analytic West-Germanic language, that descended from Dutch settlers \cite{roberge2002afrikaans}.

\section{Related Work}

Machine translation research for Southern African languages is scarce  \cite{martinus2019focus,abbott2018towards,wilken2012developing,mckellar2014an}. Much of the research suffers from the problems described above. For example, Van Niekerk  report BLEU scores as high as 37.9 for English-to-Northern Sotho translation and 71 for English-to-Afrikaans translation, but neither the datasets nor the code are published \cite{van2014exploring}. Wilken \textit{et.  al.} use publicly-available parallel datasets, but reports using monolingual datasets which were not released \cite{wilken2012developing}.

\section{Data}
\label{data}

The Autshumato parallel corpora are aligned parallel corpora of South African governmental data \cite{groenewald2009introducing}, for English to Afrikaans, isiZulu, Northern Sotho, Setswana, and Xitsonga.\footnote{Available online at: \url{https://repo.sadilar.org/handle/20.500.12185/404}} 

\begin{table}[h]
\begin{center}
\begin{tabular}{|l|l|l|l|l|l|}
\hline \bf  & \bf Afrikaans & \bf isiZulu & \bf Northern Sotho & \bf Setswana & \bf Xitsonga \\ \hline
Sentences & 53 172 & 26 728 & 30 777 & 123 868 & 193 587\\
\hline
\end{tabular}
\end{center}
\caption{\label{sentence-table} Number of parallel sentences for English-to-Target language}
\end{table}

Table \ref{sentence-table} highlights how low resourced these datasets are. The datasets had many duplicates which we removed to avoid data leakage between training and test sets.

A newly proposed Autshumato evaluation set for machine translation of South African languages has been released \cite{mckellar2017autshumato}. The evaluation set consists of 500 sentences translated separately by four different professional human translators for each of the 11 official South African languages. We evaluated our models on this benchmark.

\section{Architectures}
\label{architectures}
Limited work has been done using NMT techniques for African languages. We trained translation models for two popular NMT architectures, namely, ConvS2S \cite{gehring2017convolutional} and Transformer \cite{vaswani2017attention}\footnote{The hyper-parameters used for each architecture are available online here: \url{https://github.com/LauraMartinus/ukuxhumana} }.

\section{Results}

The BLEU scores for each model on English-to-Target language are presented in Table~\ref{bleu-table}. The Transformer model outperformed the ConvS2S model for all languages. These results serve as initial baseline results for the given languages on the evaluation set.  

\begin{table}[h]
\begin{center}
\begin{tabular}{|l|l|l|l|l|l|}
\hline \bf Model & \bf Afrikaans & \bf isiZulu & \bf Northern Sotho & \bf Setswana & \bf Xitsonga \\ \hline
ConvS2S & 12.30 & 0.52 & 7.41 & 10.31 & 10.73\\
Transformer & \textbf{20.60} & \textbf{1.34} & \textbf{10.94} & \textbf{15.60} & \textbf{17.98}\\
\hline
\end{tabular}
\end{center}
\caption{\label{bleu-table} BLEU Scores for English-to-Target language on evaluation dataset}
\end{table}

We supply a few qualitative sample translations in Table~\ref{sent-result-table-zulu} and Table~\ref{sent-result-table-xitsonga}, as well as multi-head attention visualisations, to demonstrate the performance of the models. 

We notice that the performance of techniques on a specific target language is related to corpus size and morphological typology of the language. Afrikaans is not agglutinative, thus despite having less than half the number of sentences as Xitsonga and Setswana, the model achieves better performance than the other languages. Xitsonga and Setswana are both agglutinative, but have much more data, so their models achieve higher performance than Northern Sotho or isiZulu. The models for isiZulu achieved the worst performance. We attribute the bad performance to the morphological complexity of the language (discussed in Section~\ref{languages}), the size of the dataset and poor quality of the data.

\begin{table*}

\caption{\textbf{English to isiZulu Translations}: We show the reference translation, translation by the Transformer model, and translation back to English performed by an isiZulu speaker.}
  \label{sent-result-table-zulu}
  \centering
  \small
  \begin{tabular}{|p{3cm}|p{12cm}|}
  
    \hline
    
    Source  & \textcolor{purple}{Note that the funds will be held against the Vote of the Provincial Treasury pending disbursement to the SMME Fund .}\\
    Target & Lemali izohlala emnyangweni wezimali .\\\hline
    & \\
    Transformer & Qaphela ukuthi izimali zizobanjwa kweVME esifundazweni saseTreasury zezifo ezithunyelwa ku-MSE .\\
    Back Translation &  \textcolor{purple}{Be aware that the money will be held by VME with facilities of Treasury with diseases sent to MSE .}\\\hline
    
  \end{tabular}
\end{table*}

\begin{table*}
\caption{\textbf{English to Xitsonga Translations}: We show the reference translation, translation by the Transformer model, and translation back to English performed by a Xitsonga speaker.}
  \label{sent-result-table-xitsonga}
  \centering
  \small
  \begin{tabular}{|p{3cm}|p{12cm}|}
    \hline
    
    Source & \textcolor{blue}{we are concerned that unemployment and poverty persist despite the economic growth experienced in the past 10 years .}\\
    Target & hi na swivilelo leswaku mpfumaleko wa mitirho na vusweti swi ya emahlweni hambileswi ku nga va na ku kula ka ikhonomi eka malembe ya 10 lawa ya hundzeke .\\\hline
    Transformer & hi na swivilelo leswaku mpfumaleko wa mitirho na vusweti swi ya emahlweni hambileswi ku nga va na ku kula ka ikhonomi eka malembe ya 10 lawa ya hundzeke .\\ 
    Back Translation & \textcolor{blue}{We have concerns that there is still lack of jobs and poverty even though there has been economic growth in the past 10 years.}\\\hline
   
  \end{tabular}
\end{table*}

\begin{figure*}[h!]
\centering
\begin{subfigure}{0.5\textwidth}
  \centering\captionsetup{width=.8\linewidth}
  \includegraphics[width=.92\linewidth]{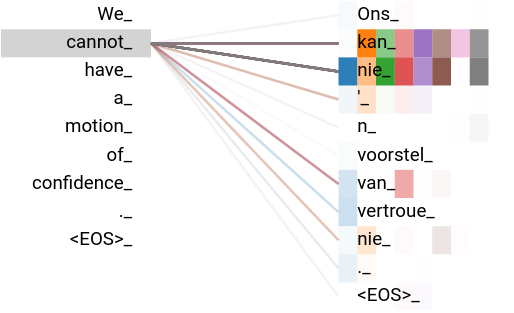}
  \caption{Visualization of multi-head attention for Layer 1 for the word ``cannot''. The coloured bars are individual attention heads. The word ``cannot'' is translated to ``kan nie ... nie'' where the second negative ``nie'' occurs at the end of the sentence.}
  \label{fig:sub1}
\end{subfigure}%
\begin{subfigure}{0.5\textwidth}
  \centering\captionsetup{width=.8\linewidth}
  \includegraphics[width=.9\linewidth]{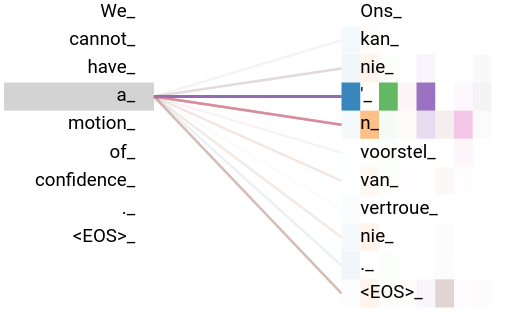}
  \caption{Visualization of multi-head attention for Layer 2 for the word ``a''. The coloured bars are individual attention heads. The word ``a'' is translated to ``'n'', as is successfully captured by the attention mechanism.}
  \label{fig:sub2}
\end{subfigure}
\caption{Visualizations of multi-head attention for an English sentence translated to Afrikaans using the Transformer model.}
\label{fig:afrikaansattention}
\end{figure*}

\section{Conclusion and Future Work}
By publishing the code, datasets and baseline results on a discoverable platform, for machine translation of the above languages and evaluating them on a publicly-available evaluation set, we begin to address the problems of discoverability, reproducibility, and comparability in the research of machine translation of African languages. Future work will include results for other Southern African languages so we can provide baselines for all official languages in South Africa, by using unsupervised and transfer-learning NMT techniques.

\section*{Acknowledgements}
The authors would like to thank Mbongiseni Ncube, Reinhard Cromhout, and Vongani Maluleke for assisting us with the back-translations. Research supported with Cloud TPUs from Google's TensorFlow Research Cloud (TFRC).
%
%
\blfootnote{
    %
    %
    %
    %
     \hspace{-0.65cm}  
     This work is licensed under a Creative Commons 
     Attribution 4.0 International Licence.
     Licence details:
     \url{http://creativecommons.org/licenses/by/4.0/}.
    %
    %
}

\bibliographystyle{acl.bst}
\bibliography{ref}
\end{document}